\documentclass[runningheads]{llncs}

\usepackage{eccv}

\usepackage{eccvabbrv}

\usepackage{graphicx}
\usepackage{booktabs}

\usepackage[accsupp]{axessibility}  

\usepackage{hyperref}

\usepackage{orcidlink}

\usepackage{amsmath}
\usepackage{tikz}
\usepackage{bm}
\usepackage[misc]{ifsym}

\newcommand{\probP}{\mathbb{P}}
\newcommand{\probM}{\mathbb{M}}

\newcommand\blfootnote[1]{%
  \begingroup
  \renewcommand\thefootnote{}\footnote{#1}%
  \addtocounter{footnote}{-1}%
  \endgroup
}

\begin{document}

\title{Efficient Diffusion Transformer with Step-wise Dynamic Attention Mediators} 


\author{Yifan Pu\inst{1}$^*$\orcidlink{0000-0002-0404-1737} \and
Zhuofan Xia\inst{1}$^*$\orcidlink{0009-0001-7965-364X} \and
Jiayi Guo\inst{1}$^*$\orcidlink{0009-0005-7004-939X} \and
Dongchen Han\inst{1}\orcidlink{0009-0009-3431-6189} \and \\
Qixiu Li\inst{1}\orcidlink{0009-0002-4866-6920} \and
Duo Li\inst{1}\orcidlink{0009-0008-3524-1935} \and
Yuhui Yuan\inst{2}\orcidlink{0000-0002-8345-4205} \and
Ji Li\inst{2}\orcidlink{0000-0003-4699-084X} \and
Yizeng Han\inst{1}\orcidlink{0000-0001-5706-8784} \and \\
Shiji Song\inst{1}\orcidlink{0000-0001-7361-9283} \and
Gao Huang\inst{1}\textsuperscript{(\Letter)}\orcidlink{0000-0002-7251-0988} \and
Xiu Li\inst{1}\textsuperscript{(\Letter)}\orcidlink{0000-0003-0403-1923}
}

\authorrunning{Y. Pu et al.}

\institute{Tsinghua University, Beijing 100084, China\\
\email{\{puyf23, xzf23, guo-jy20\}@mails.tsinghua.edu.cn}\blfootnote{$*$ Equal contribution. \Letter~Corresponding authors.}\\
\email{\{shijis, gaohuang\}@tsinghua.edu.cn} \and
Microsoft Research Asia \\
\email{\{yuhui.yuan, ji.li\}@microsoft.com}}

\maketitle

\begin{abstract}
    This paper identifies significant redundancy in the query-key interactions within self-attention mechanisms of diffusion transformer models, particularly during the early stages of denoising diffusion steps.
    In response to this observation, we present a novel diffusion transformer framework incorporating an additional set of mediator tokens to engage with queries and keys separately. By modulating the number of mediator tokens during the denoising generation phases, our model initiates the denoising process with a precise, non-ambiguous stage and gradually transitions to a phase enriched with detail.
    Concurrently, integrating mediator tokens simplifies the attention module's complexity to a linear scale, enhancing the efficiency of global attention processes. Additionally, we propose a time-step dynamic mediator token adjustment mechanism that further decreases the required computational FLOPs for generation, simultaneously facilitating the generation of high-quality images within the constraints of varied inference budgets.
    Extensive experiments demonstrate that the proposed method can improve the generated image quality while also reducing the inference cost of diffusion transformers. When integrated with the recent work SiT, our method achieves a state-of-the-art FID score of 2.01. The source code is available at \url{https://github.com/LeapLabTHU/Attention-Mediators}.
  
\end{abstract}
\section{Introduction}
\label{sec:intro}

Exhibiting unprecedented capabilities in the fields of language processing~\cite{devlin2018bert,brown2020language,raffel2020exploring,touvron2023llama,achiam2023gpt} and visual recognition~\cite{liu2021swin,oquab2024dinov,fang2023eva,kirillov2023segment,radford2021learning}, Transformers~\cite{vaswani_attention} have recently achieved remarkable performance in visual generation as backbones in diffusion models~\cite{peebles2023scalable, videoworldsimulators2024}. The inherent simplicity, effectiveness, and scalability of these Diffusion Transformers (DiTs) position themselves as appealing alternatives to previously prominent U-Net structures~\cite{ronnenberger15, rombach2022high, saharia2205photorealistic, ramesh2021zero}, promoting the emergence of high-resolution and high-quality image/video generation applications, such as Stable Diffusion V3~\cite{sd3}, Pixart-$\alpha/\Sigma/\delta$~\cite{chen2024pixartsigma, chen2024pixartdelta, chen2023pixartalpha}, Hunyuan-DiT~\cite{li2024hunyuan} and Sora~\cite{videoworldsimulators2024}.

Despite the rapid progress of Diffusion Transformers, widespread criticism has arisen due to their substantial consumption of computing resources and the associated inference time overhead~\cite{chen2023pixartalpha,mo2024dit,xue2024sa} resulting from the global attention mechanism. This obstacle impedes the practical deployment of Diffusion Transformers for large-scale client usage, particularly when dealing with high-resolution images~\cite{chen2023pixartalpha,lu2024fit} and relatively long videos~\cite{lu2023vdt,ma2024latte}. While several works~\cite{zheng2023fast, crowson2024scalable, gao2023masked} have been proposed to accelerate the attention process in visual recognition tasks, this topic remains largely unexplored in the realm of visual generation. Therefore, it is crucial to develop an efficient Diffusion Transformer to address high resource consumption concerns and enhance overall usability.
 
In this paper, we expedite the diffusion generation process by leveraging the inherent structural redundancy~\cite{xia2023budgeted,michel2019sixteen,zhang2021enlivening,song2021dynamic} in Diffusion Transformers across different denoising time steps. We start by identifying the redundancies in the query-key interaction process during the self-attention operation at each layer in Transformer diffusers. To analyze quantitatively, we design a Jensen–Shannon divergence-based metric to measure the query-key interaction redundancy, \ie, comparing the attention distribution similarities among each query. We come up with two key findings: (1) Extensive query-key redundancy is evident in all of the self-attention layers, indicating many tokens would be homogeneous after self-attention; (2) The redundancy is particularly pronounced in the initial steps while gradually diminishing in the subsequent steps as denoising goes on, suggesting the fully one-to-one attention in the early steps be dispensable.

To fully take advantage of this redundancy, we introduce an extra set of tokens in the conventional self-attention layers, dubbed \textbf{attention mediators}, to streamline the interaction process between queries and keys, condensing the actual interactions in the attention between queries and keys. To be specific, the number of mediator tokens is set lower than that of queries and keys, \eg, less than 10\% of the original tokens. These mediator tokens first aggregate the information from keys with softmax attention, forming packed representations. Then, the compressed information is propagated to queries in another softmax attention as the final output. The abbreviated mediators bottleneck the attention and hence confine its redundancy, further reducing the computation cost via interchanging the attention computation order.

In addition to attention mediators, the redundancy variations across time steps elicit a new dynamic strategy for adjusting the number of mediator tokens at different time steps. Specifically, during the early steps where the redundancy is prominent, we utilize a smaller number of mediator tokens to reduce similar information aggregation effectively. When redundancy gradually diminishes during the later steps, we dynamically increase the number of mediator tokens to generate more detailed and diversified features. In practice, the schedule of switching mediators is determined by the samples' latent distance between each pair of adjacent denoising steps. This dynamic strategy maintains mediator token efficiency while enhancing generation quality and diversity.

We evaluated our proposed method using the very recent SiT~\cite{ma2024sit} model. Extensive experimental results demonstrate that our approach achieves superior generation quality (as indicated by a lower FID~\cite{heusel2017gans}) and reduces computational complexity (measured in FLOPs) during generation. When combined with the SiT-XL/2 model, our method achieves a state-of-the-art FID score.

\section{Related Works}
\label{sec:related}

\subsection{Diffusion Transformers}

Recent advancements in diffusion models~\cite{dhariwal2021diffusion, ramesh2022, ho2020denoising, liu2024scoft, guo2024smooth} have typically utilized the U-Net architecture~\cite{ronnenberger15}. However, a growing body of research~\cite{yang2022your, peebles2023scalable, bao2023all} has begun to explore the potential of employing the Vision Transformer (ViT)~\cite{vit20} as an alternative backbone for such models. U-ViT~\cite{bao2023all} interprets various inputs (\eg, time, conditions, and noisy image patches) as tokens while drawing inspiration from U-Net to implement skip connections between the model's shallow and deep layers. DiT~\cite{peebles2023scalable} demonstrates the scalability of ViT for diffusion models, surpassing the performance of U-Net-based diffusion models on ImageNet. Building upon DiT, SiT~\cite{ma2024sit} introduces an interpolant framework, moving from discrete to continuous time and exploring various diffusion coefficients, thereby achieving superior results. MaskDiT~\cite{zheng2023fast} pioneers the use of masked training to reduce the computational expense of training diffusion models. MDT~\cite{gao2023masked} additionally proposes a masked latent modeling technique, and MDTv2 further refines this approach with a more efficient macro network architecture and training strategy, improving the FID and accelerating the learning process. HDiT~\cite{crowson2024scalable} leverages transformers to devise a high-resolution training methodology that scales linearly with pixel count. FiT~\cite{zheng2023fast} conceptualizes images as sequences of dynamically sized tokens to generate images, facilitating image generation at varying resolutions and aspect ratios. These investigations confirm that transformer-based models are effective in visual generation tasks and can be scalable. Although these works have demonstrated the effectiveness of transformers in diffusion models and have further improved the FID or training speed by optimizing the diffusion structure or learning strategies, the inner design structure of the Diffusion Transformer backbone is still not well explored.

\subsection{Attention with Linear Complexity}

One line of works achieves linear computational complexity by restricting receptive fields, including Shifted-window attention~\cite{liu2021swin}, Neighborhood Attention~\cite{hassani2023neighborhood}. These works bring locality back into the vision transformer architecture, while the global context awareness is somewhat affected. In contrast to the idea of restricting receptive fields, another line of researcgh directly uses linear attention to address the computational challenge by reducing computation complexity. The pioneer work~\cite{linear_attn} discards the Softmax function and replaces it with a mapping function $\phi$ applied to $Q$ and $K$, thereby reducing the computation complexity to $\mathcal{O}(N)$. However, such approximations led to substantial performance degradation. To tackle this issue, Efficient Attention~\cite{shen2021efficient} applies the Softmax function to both $Q$ and $K$. SOFT~\cite{lu2021soft} and Nyströmformer~\cite{xiong2021nystromformer} employ matrix decomposition to further approximate Softmax operation. Castling-ViT~\cite{you2022castling} uses Softmax attention as an auxiliary training tool and fully employs linear attention during inference. FLatten Transformer~\cite{han2023flatten} proposes a focused function and adopts depthwise convolution to promote feature diversity limited by linear operations.

Furthermore, Agent Attention~\cite{han2023agent} and Anchored Stripe Attention~\cite{li2023efficient} introduce another group of tokens as the bridge between queries and keys, which is equivalent to linear attention, achieving favorable performance on recognition tasks and low-level visions, respectively. In this paper, we build our work upon this architecture and comprehend the extra group of tokens as semantically compressed information to guide the diffusion process to generate images.

\subsection{Dynamic Neural Networks}

In contrast to static models, which have fixed computational graphs and parameters at the inference stage, dynamic neural networks~\cite{han2021dynamic,wang2023computation} can adapt their structures or parameters to different inputs, leading to notable advantages in terms of performance~\cite{tang2024mind}, adaptiveness~\cite{yang2023hundreds, guo2023zero}, computational efficiency~\cite{yang2023boosting,wang2022efficient}, and representational power~\cite{pu2023adaptive}. Dynamic networks are typically categorized into three types: sample-wise~\cite{huang2017multi,wang2021not,han2022learning, han2023dynamic,pu2023fine, wang2024gra,xia2024gsva}, spatial-wise~\cite{wang2021glancing,huang2022glance,han2022latency,han2023latency,han2021spatially,xia2022vision,xia2023dat++,pan2023slide}, and temporal-wise~\cite{hansen2019neural,wang2021adaptive}. Since the breakthrough query-based visual recognition model DETR~\cite{carion20}, a new query-based dynamic network has begun to develop~\cite{pu2024rank}.
In this work, we introduce a novel temporal-wise dynamic approach. Contrary to the former works, which study the dynamic mechanism along the video time dimension~\cite{wang2021adaptive,he2023camouflaged,wang2022adafocusv3}, we explore the redundancy across the diffusion-denoising time steps in this paper. We dynamically change the number of mediator tokens, conditioned on the generation process of different image samples, and achieve better FID-50K results with less computational complexity.

\section{Attention Redundancies Along Denoising Steps}
\label{sec:redun}

In this section, we examine redundancies in conventional self-attention operations. Initially, we provide a brief overview of attention computation in Transformer architectures. Subsequently, we introduce a quantitative metric designed to analyze redundancies in query-key interactions. Our findings reveal that significant redundancies exist in Diffusion Transformers, and the extent of this redundancy decreases as the denoising procedure progresses.

\subsection{Background of Attention}
\label{sec:redun_attn}

We first revisit the attention mechanism~\cite{vaswani_attention} in Diffusion Transformers~\cite{peebles2023scalable,ma2024sit}. The latent Diffusion Transformer takes a latent token sequence $\bm{z}_{l-1}\!\in\!\mathbb{R}^{N\times{}C}$ from the previous layer $l-1$ as input ($N$ is the token number and $C$ is the hidden dimension), then projects it into the query, key, and value sequences with three linear projection layers, denoted as $\mathbf{W_q},\mathbf{W_k},\mathbf{W_v}\!\in\!\mathbb{R}^{C\times{}C}$ (bias omitted):
\begin{equation}
\bm{q}=\bm{z}_{l-1}\mathbf{W_q},\hspace{4pt}\bm{k}=\bm{z}_{l-1}\mathbf{W_k},\hspace{4pt}\bm{v}=\bm{z}_{l-1}\mathbf{W_v}.
\label{eq:attn_qkv}
\end{equation}
Then $\bm{q},\bm{k},\bm{v}\!\in\!\mathbb{R}^{N\times{}C}$ are divided into $M$ heads $\bm{q}^{(m)},\bm{k}^{(m)},\bm{v}^{(m)}\!\in\!\mathbb{R}^{N\times{}d}$ in terms of channel $C$, with head dimension of $d\!=\!C/M$. Within each head, the similarity of each query $\bm{q}^{(m)}$ and key $\bm{k}^{(m)}$ is computed as:
\begin{equation}
\mathbf{A}^{(m)}=\text{Softmax}\left(\bm{q}^{(m)}\bm{k}^{(m)\top}/\sqrt{d}\right),
\label{eq:attn_softmax}
\end{equation}
where the attention map $\mathbf{A}^{(m)}$ is an $N\!\times{}\!N$ matrix containing elements in the range $[0,1]$, and the sum of each row is normalized to 1. The attention mechanism reweights the value sequence according to the attention map, $\bm{h}^{(m)}\!=\!\mathbf{A}^{(m)}\bm{v}^{(m)}$, to dynamically adjust the outputs based on the dependency of each token in the inputs. In the end, each head of the reweighted representation is concatenated together to produce the final output of this layer $l$, written as:
\begin{equation}
\bm{z}_{l}=\text{Concat}\left(\bm{h}^{(1)},\bm{h}^{(2)},\ldots,\bm{h}^{(M)}\right)\mathbf{W_O},
\label{eq:attn_out}
\end{equation}
where $\mathbf{W_O}\!\in\!\mathbb{R}^{C\times{}C}$ (bias omitted) is a linear projection layer to promote interaction between different heads in the multi-head attention layer.

We view each row of $\mathbf{A}^{(m)}$ in \cref{eq:attn_softmax} as a probabilistic distribution between one query and all the keys, \eg, the $i$-th row $\mathbf{A}_{i}^{(m)}\!\in\!\mathbb{R}^{1\times{}N}$ depicts how the $N$ key tokens contribute to the output of the $i$-th query token, on the $m$-th attention head. Since the output of $i$-th token $\bm{h}_i^{(m)}\!=\!\mathbf{A}_{i}^{(m)}\left(\bm{z}_{l-1}\mathbf{W_v}\right)^{(m)}$ only distinguishes other tokens by the distribution $\mathbf{A}_{i}^{(m)}$, the feature diversity in the output sequence of the attention is determined by this distribution. If different queries $\bm{q}_{i_1}$ and $\bm{q}_{i_2}$ ($i_1\!\ne{}\!i_2$) share similar probabilistic distributions over keys, \ie, $\mathcal{D}\left(\mathbf{A}_{i_1}^{(m)},\mathbf{A}_{i_2}^{(m)}\right)\!\approx{}\!0$ for some distribution similarity metric $\mathcal{D}(\cdot,\cdot)$, the output $\bm{h}_{i_1}^{(m)}$ and $\bm{h}_{i_2}^{(m)}$ would be rather close, leading to redundant representations and a lack of spatial diversity in the diffusion noise prediction process.

\subsection{Jensen-Shannon Divergence as A Redundancy Metric}
\label{sec:jsd}

We adopt Jensen-Shannon Divergence (JSD) as the redundancy metric $\mathcal{D}$ to study the spatial redundancy in attention on latent tokens quantitatively. JSD is a symmetric divergence that combines two Kullback–Leibler Divergence (KLD). Given two probabilistic distributions $\probP_1(X)$ and $\probP_2(X)$ in which $X$ is a discrete random variable with $K$ possible values, the KLD is defined as 
\begin{equation}
\mathcal{D}_\text{KL}\left(\probP_1{}\Vert{}\probP_2{}\right)=\sum_{k=1}^{K}\probP_1(X\!=\!k)\left[\ln{}\probP_1(X\!=\!k)-\ln{}\probP_2(X\!=\!k)\right].
\label{eq:kld}
\end{equation}
Then the JSD is defined with a mixture distribution $\probM{}\!=\!\frac{1}{2}\left(\probP_1{}+\probP_2{}\right)$, by averaging the KLD of $\probP_1{}$ from $\probM{}$ and the KLD of $\probP_2{}$ from $\probM{}$, written as
\begin{equation}
\mathcal{D}_\text{JS}(\probP_1{}\Vert{}\probP_2{})=\dfrac{1}{2}\left[\mathcal{D}_\text{KL}(\probP_1{}\Vert{}\probM{})+\mathcal{D}_\text{KL}(\probP_2{}\Vert{}\probM{})\right].
\label{eq:jsd}
\end{equation}
The JSD is symmetric and bounded in that $\mathcal{D}_\text{JS}(\probP_1{}\Vert{}\probP_2{})=0$ when $\probP_1{}$ and $\probP_2{}$ are identical, and $\mathcal{D}_\text{JS}(\probP_1{}\Vert{}\probP_2{})\!\to{}\!\ln{}2$ when the support of $\probP_1{}$ and $\probP_2{}$ are disjoint. JSD decreases as two distributions are closer and increases vice versa.

For the query token sequence, we compare the attention distribution by each pair of queries using Jensen-Shannon Divergence and then accumulate the divergence to each query token as the final redundancy score metric, which we define as follows for the $l$-th layer in the Diffusion Transformers:
\begin{equation}
S_l=\dfrac{2}{MN(N\!-\!1)}\sum_{m=1}^{M}\sum_{i_1=1}^{N-1}\sum_{i_2=i_1+1}^{N}\mathcal{D}_\text{JS}\left(\mathbf{A}_{i_1}^{(m)},\mathbf{A}_{i_2}^{(m)}\right).
\label{eq:score}
\end{equation}

\noindent
This score computes the JSD of every attention distribution pair in the latent token sequence and reduces over $\frac{N(N\!-\!1)}{2}$ pairs and $M$ attention heads. A \emph{high} $S_l$ means that the averaged attention maps among the tokens are in \emph{low} similarity in the $l$-th layer, indicating a \emph{low} spatial redundancy. On the contrary, a low $S_l$ means the redundancy in the $l$-th layer is relatively high.

\begin{figure}[t]
  \begin{minipage}[t]{0.49\linewidth}
  \centering
  \includegraphics[width=0.99\linewidth]{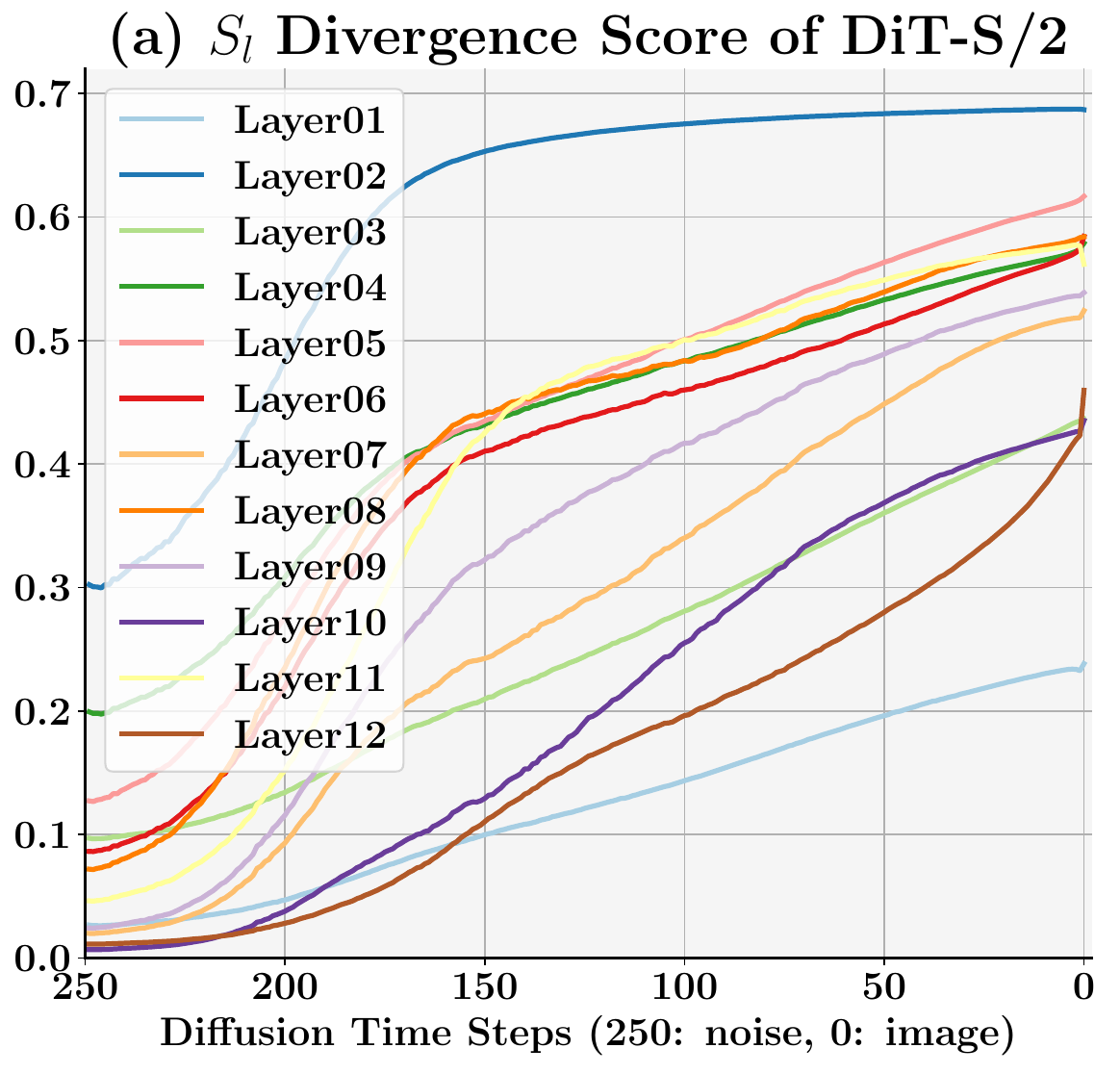}
  \end{minipage}
  \begin{minipage}[t]{0.49\linewidth}
  \centering
  \includegraphics[width=0.99\linewidth]{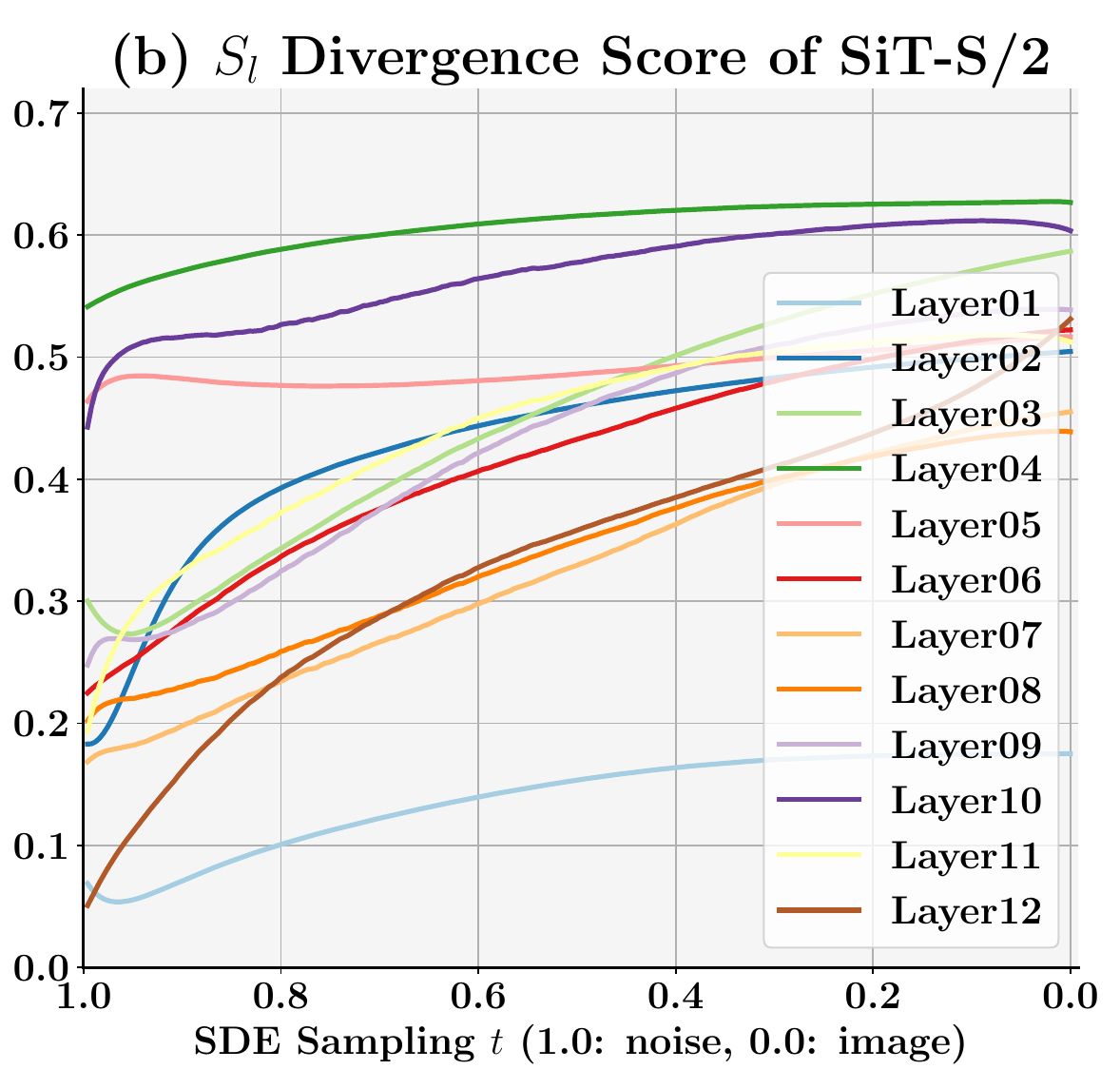}
  \end{minipage}
  \caption{(a) shows the JSD-based redundancy score defined in \cref{sec:jsd} evaluated on DiT-S/2 model along with diffusion time steps. The score is computed over 32 samples and averaged by different attention heads in every layer. (b) shows the same redundancy score of all the 12 layers of SiT-S/2 model with the SDE sampler.}
  \label{fig:redu}
\end{figure}

\subsection{Redundancies Along Time Steps}
\label{sec:redun_alt}

We measure the $S_l$ of both the DiT-S/2~\cite{peebles2023scalable} and SiT-S/2 model~\cite{ma2024sit}. We randomly sample 512 images with a pretrained model and record the $S_l$ of all the diffusion transformer layers and all the denoising time steps (all the SDE sampling $t$ in SiT). The results for DiT-S/2 and SiT-S/2 are illustrated in \cref{fig:redu}(a) and \cref{fig:redu}(b), respectively. Notably, the redundancy of self-attention is \emph{inversely} proportional to $S_l$. As a result, we get two observations from \cref{fig:redu}. First, massive query-key redundancy exists in the attention operation of the diffusion transformers. For example, in some layers (\eg layer 10 in DiT-S/2), the inner-query distance is nearly zero in the first several time steps, implying that almost all the queries are akin and redundant. The second observation is that redundancy gradually decreases as the denoising process continues. It is implied that the queries become more diverse in the latter denoising steps. 

Based on the above phenomenon, we design mediator tokens that interact with query and key tokens separately, thus compressing the excessive attention between queries and keys. The number of mediator tokens can be adjusted in different time steps, thus adapting the different degrees of redundancy inside different phases of the denoising process. We present the detailed explanation of our method in the following section.

\section{Efficient DiTs with Attention Mediators}
\label{sec:method}

In this section, we introduce the attention mediator mechanism to leverage the redundancy efficiently in \cref{sec:attn_med}, building up a dynamic architecture of Diffusion Transformer. To further boost the efficiency of Dynamic Diffusion Transformers, we devise an algorithm in \cref{sec:time_step_dynamic} to speed up the sampling process and fit the computational budgets via dynamically adjusting mediator tokens.

\subsection{Attention Mediators}
\label{sec:attn_med}

We present the attention mediators to regulate the attention between every two query and key pairs. The high-level idea of attention mediators is to use an additional group of tokens to compress the interaction between the queries and keys. The additional tokens, which we name it as mediator tokens, usually have a smaller number than queries or keys, serving as a condensed supervisor over the attention interaction. We present the detail as follows.

In each head of the multi-head attention module, besides the query $\bm{q}^{(m)}$, key $\bm{k}^{(m)}$, and value $\bm{v}^{(m)}$ tokens, we introduce a set of mediator tokens $\bm{t}^{(m)} \in \mathbb{R}^{n \times d}$, where $n$ is the mediator token length and $n \ll N$. The mediator tokens first interact with the key tokens to get the intermediate result $\bm{v}_{\text{med}}^{(m)}$:

\begin{equation}
\bm{v}_{\text{med}}^{(m)}=\text{Softmax}\left(\bm{t}^{(m)}\bm{k}^{(m)\top}/\sqrt{d}\right)\bm{v}^{(m)},
\label{eq:tk}
\end{equation}

\noindent
where $\bm{v}_{\text{med}}^{(m)} \in \mathbb{R}^{n \times d}$. Then the mediator token interacts with the query tokens and extracts the results from the intermediate result $\bm{v}_{\text{med}}^{(m)}$:

\begin{equation}
\bm{h}^{(m)}=\text{Softmax}\left(\bm{q}^{(m)}\bm{t}^{(m)\top}/\sqrt{d}\right) \bm{v}_{\text{med}}^{(m)}.
\label{eq:qt}
\end{equation}

\noindent
In this way, a condensed set of mediator tokens interacts with the queries and keys separately, avoiding redundancy when they interact indirectly.

The mediator tokens are obtained by adaptively pooling the query tokens into a small number of tokens. Considering the noise predicted by the transformer has spatial structured information, we first reshape the query tokens into the latent image shape $\mathbb{R}^{H \times W \times d}$ and then pool it in the spatial dimensions to get $\mathbb{R}^{h \times w \times d}$. The pooled queries are finally reshaped to the mediator tokens $\bm{t}^{(n)} \in \mathbb{R}^{n \times d}$, where $n \ll N$ because $(h \times w) \ll (H \times W)$.

\subsection{Complexity Analysis}

It is noteworthy that by incorporating an additional, compact set of tokens, we achieve a reduction in redundancy within the attention mechanism. Simultaneously, the computational complexity inherent to the attention operation is diminished. We provide the subsequent analysis.

We begin by mixing and combining \cref{eq:tk} and \cref{eq:qt} to formulate the final output of self-attention with mediator tokens:
\begin{equation}
\bm{h}^{(m)}=\underbrace{\text{Softmax}\left(\bm{q}^{(m)}\bm{t}^{(m)\top}/\sqrt{d}\right)\underbrace{\text{Softmax}\left(\bm{t}^{(m)}\bm{k}^{(m)\top}/\sqrt{d}\right)\bm{v}^{(m)}}_{\text{Step 1: }\mathbb{R}^{n\times{}N}\bm{\cdot}\ \mathbb{R}^{N\times{}d}\ \to{}\ \mathcal{O}(Nnd)}}_{\text{Step 2: }\mathbb{R}^{N\times{}n}\bm{\cdot}\ \mathbb{R}^{n\times{}d}\ \to{}\ \mathcal{O}(Nnd)}.
\label{eq:med_attn}
\end{equation}
Since queries $\bm{q}^{(m)}$ and keys $\bm{k}^{(m)}$ are decoupled by the mediators, we can interchange the computation order of the queries, keys and values in attention. Unlike previous vanilla self-attention that firstly computes $\bm{q}^{(m)}$ and $\bm{k}^{(m)}$, we first aggregate values $\bm{v}^{(m)}$ with precomputed $\bm{A}_{\text{tk}}^{(m)}=\text{Softmax}\left(\bm{t}^{(m)}\bm{k}^{(m)\top}/\sqrt{d}\right)$, as shown in Step 1 of \cref{eq:med_attn}. The complexity of step 1 in multiplying an $n\times{}N$ matrix and an $N\times{}d$ matrix is $\mathcal{O}(Nnd)$, as well as computing $\bm{A}_{\text{tk}}^{(m)}$, which involves multiplying an $n\times{}d$ matrix and an $N\times{}d$ matrix. Thus, the overall complexity of Step 1 is no more than $2Nnd$, also controlled by $\mathcal{O}(Nnd)$. The result of Step 1 has the shape of $\mathbb{R}^{n\times{}d}$, therefore the information propagation to queries of step 2 with $\bm{A}_{\text{qt}}^{(m)}=\text{Softmax}\left(\bm{q}^{(m)}\bm{t}^{(m)\top}/\sqrt{d}\right)$ is also an $\mathcal{O}(Nnd)$ complex operation.

To summarize, both Steps 1 and 2 in \cref{eq:med_attn} have $\mathcal{O}(Nnd)$ complexity, with $N$ latent tokens, $n$ mediator tokens, and $d$ feature dimensions in each attention head. The proposed attention module achieves linear complexity relative to $N$, $n$, and $d$. Summing all heads together, the proposed mediator attention has an  $\mathcal{O}(nNC)$ complexity. Compared with the vanilla self-attention, which directly multiplies queries and keys together to aggregate values and get $\mathcal{O}(N^2C)$ complexity, our method significantly reduces computational demands, given that the mediator token count $n$, is significantly less than the image token count $N$. To compensate the potential loss of feature diversity in linear complexity attention, we adopt a depthwise convolution following  Flatten transformer~\cite{han2023flatten}.

\subsection{Time Step-wise Mediator Adjusting}
\label{sec:time_step_dynamic}

\cref{fig:redu} illustrates the variation in attention redundancy across different diffusion denoising time steps, revealing a gradual decrease in redundancy throughout the process. Understanding the attention mediator tokens as a means of compressing tokens between query and value tokens, we exploit this phenomenon, as shown in \cref{fig:redu}, to dynamically adjust the number of mediator tokens, increasing them from loss to more along the diffusion denoising steps.

Given the variability of the denoising procedure across image samples, we introduce a sample-specific method for dynamically adjusting the number of mediator tokens. This approach allows for a customized mediator token adjustment schedule for each sample, based on its unique denoising process.

To quantify the changes in latent features between adjacent time steps, we calculate the distance between each pair of subsequent time steps, denoted as $\mathrm{\Delta}_{t} = \Vert x_{t} - x_{t+1} \Vert$, alongside recording the initial denoising difference $\mathrm{\Delta}_0 = \Vert x_{0} - x_{1} \Vert$. The denoising process begins with a Diffusion Transformer featuring a smaller number $n_1$ of mediator tokens. Upon the latent difference falling below a threshold $\rho_0$ of the initial difference $\mathrm{\Delta}_0$, we transition to a Diffusion Transformer with an increased number $n_2$ of mediator tokens. 

\begin{equation}
n_t=\left\{
\begin{aligned}
n_\text{1} &, \mathrm{\Delta}_t > \rho_0 \cdot \mathrm{\Delta}_0, \\
n_\text{2} &, \mathrm{\Delta}_t \leq \rho_0 \cdot \mathrm{\Delta}_0.
\end{aligned}
\right.
\label{eq:schedule_two}
\end{equation}

\noindent
This process is further refined by introducing additional thresholds for change, utilizing varying numbers of mediator tokens at each stage:

\begin{equation}
n_t=\left\{
\begin{aligned}
n_\text{1} &, \mathrm{\Delta}_\text{t} > \rho_0 \cdot \mathrm{\Delta}_{0}, \\
n_\text{2} &, \mathrm{\Delta}_\text{t} \leq \rho_1 \cdot \mathrm{\Delta}_{0}, \\
\vdots \\
n_\text{k} &, \mathrm{\Delta}_\text{t} \leq \rho_{\text k-1} \cdot \mathrm{\Delta}_{0}.
\end{aligned}
\right.
\label{eq:schedule_more}
\end{equation}

\section{Experiments}
\label{sec:exp}

In this section, we empirically evaluate the proposed sample-wise adaptive mediator tokens adjustment method on the state-of-the-art diffusion transformer SiT~\cite{ma2024sit}. We begin by introducing the experiment settings in \cref{sec:setup}, which include the dataset description and training hyper-parameters.  The experiment results for different numbers of mediator tokens are presented in \cref{sec:med_token_exp}.
In \cref{sec:schedule}, we show how to optimize the schedule for adjusting the mediator tokens.
Then, the effectiveness of the time step-wise mediator adjustment mechanism on larger models and higher resolutions is demonstrated in \cref{sec:main}. We also compare our method with some state-of-the-art approaches in \cref{sec:sota}. Finally, more ablation studies regarding our method and the generation visualization results are presented in \cref{sec:ablation} and \cref{sec:vis}, respectively.

\subsection{Experimental Setups}
\label{sec:setup}

Following DiT~\cite{peebles2023scalable} and SiT~\cite{ma2024sit}, we train class-conditional diffusion transformer models on the highly-competitive generative modeling benchmark ImageNet-1k~\cite{Deng2009ImageNet}. We adopt AdamW~\cite{kingma2017adam, loshchilov2019decoupled} optimizer to train all the diffusion models with no weight decay. For $256 \times 256$ image resolution models, we train them from scratch with a global batch size of $256$ for $400$K iterations. The global learning rate is set as constant $1 \times 10^{-4}$ during all training steps. We only use simple random horizontal flops data augmentation and maintain an exponential moving average (EMA) of the model weights over training with a decay of 0.9999.

\subsection{Effectiveness of Attention Mediator Tokens}
\label{sec:med_token_exp}

To verify the effectiveness of the proposed mediators, we replace the standard self-attention layers in SiT-S/2~\cite{ma2024sit} with the mediator-token ones. The experiments are conducted at a $256 \times 256$ resolution, and the images are sampled without using classifier-free guidance. \cref{tab:med_token} shows the results for different numbers of static mediator tokens, which means the token number is static across different denoising time steps. It is observed that by compressing the query-key interaction process, our method not only reduces the computational complexity in FLOPs but also achieves a higher generated image quality in FID.

\begin{table}[t]
  \caption{Effectiveness of static mediator tokens. $n$ is the mediator tokens number. }
  \label{tab:med_token}
  \centering
  \begin{tabular}{l|c|ccccc}
    \toprule
    Model & FLOPs(G) & FID ($\downarrow$) & sFID ($\downarrow$) & IS ($\uparrow$) & Precision ($\uparrow$) & Recall ($\uparrow$) \\
    \midrule
    SiT-S/2 (baseline) & 6.06 & 58.61 & 9.25 & 24.31 & 0.41 & 0.59\\
    \midrule
    + Ours ($n=4$) & \bf{5.49} & 57.67 & 10.01 & 26.66 & 0.42 & 0.56 \\
    + Ours ($n=16$) & 5.55 & 54.55 & 9.28 & 26.55 & \bf{0.43} & 0.59 \\
    + Ours ($n=64$) & 5.78 & \bf{53.57} & \bf{9.01} & \bf{27.26} & \bf{0.43} & \bf{0.61} \\
  \bottomrule
  \end{tabular}
\end{table}

\subsection{Exploring Optimized Mediator Token Adjustment Schedule}
\label{sec:schedule}

\begin{figure}
  \centering
  \includegraphics[width=0.99\linewidth]{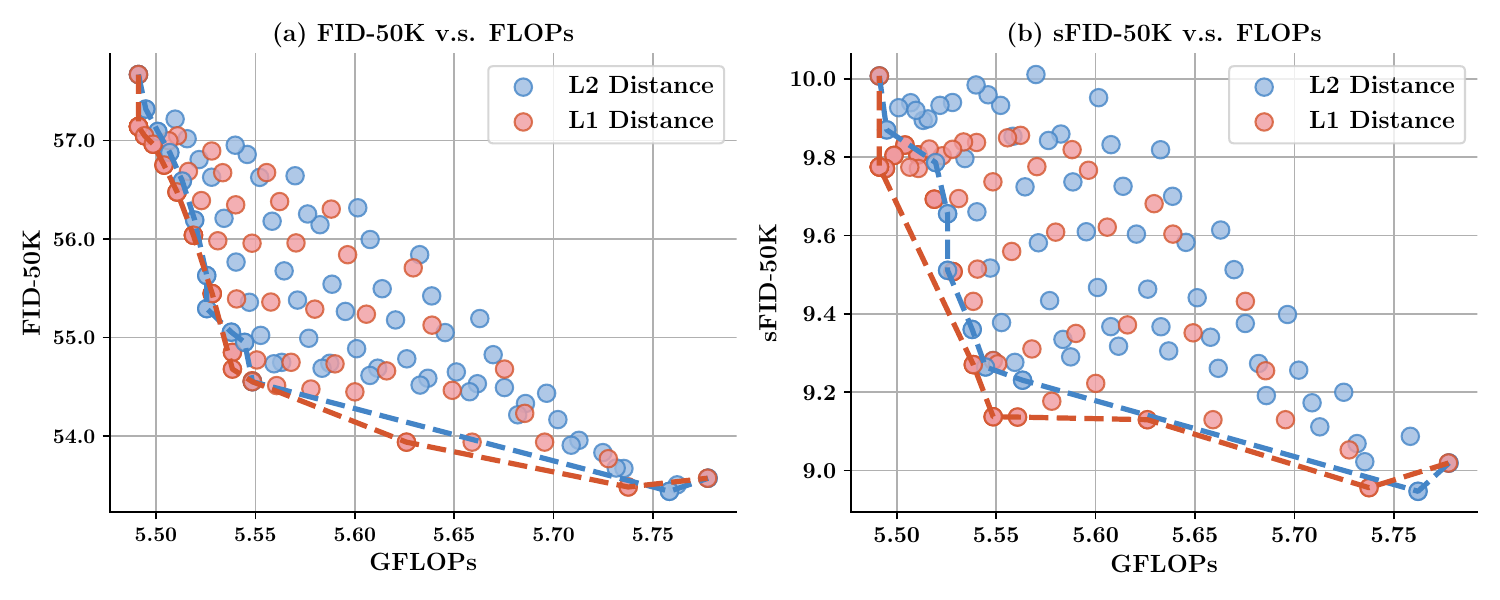}
  \caption{Ablation for optimized mediator token adjustment schedule. (a) Trade-off between FID-50K and FLOPs. (b) Trade-off between sFID-50K and FLOPs.}
  \label{fig:schedule}
\end{figure}

Since determining optimized thresholds ($\rho_i$ in \cref{eq:schedule_more}) is non-trivial, we conduct a small-scale grid search to explore reasonable mediator token number change thresholds. Specifically, we use the three models introduced in \cref{tab:med_token}. We sweep the first threshold $\rho_0$ in $\{1.0, 0.9, \cdots, 0.1, 0.0\}$, and sweep the second threshold $\rho_1$ in $\{\rho_0, \rho_0 - 0.1, \cdots, 0.1, 0.0\}$. In this way, this search space not only includes the ensemble of these three models with different numbers of mediator tokens, but also contains two-model ensembles and a single model. The choice of distance function, as described in \cref{sec:time_step_dynamic}, is also ablated between L1 and L2 distance.

The results regarding the trade-off between FID/sFID-50K and computation cost in GFLOPs are illustrated in \cref{fig:schedule}(a) and \cref{fig:schedule}(b). We plot all the results under different thresholds, along with their envelope curves. The thresholds in the envelope curves are considered optimized. We also compare the effectiveness of using L1 versus L2 distance and find that the L1 distance is the better choice.

\subsection{Main Results}\label{sec:main}

\begin{figure}[tb]
  \centering
  \includegraphics[width=1.0\linewidth]{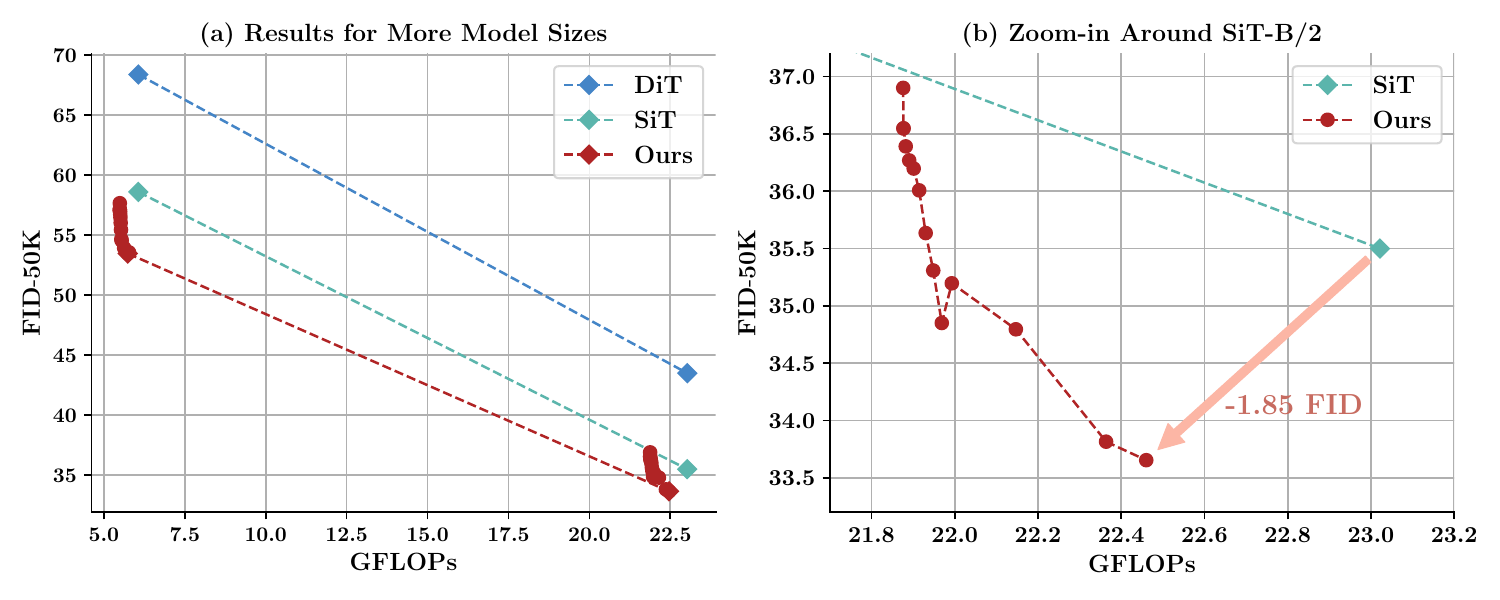}
  \caption{Main Results of the proposed method in $256\times256$ resolution. Each string of red dots is obtained by adjusting the mediator token number with optimized thresholds. (a) Comparison with DiT~\cite{peebles2023scalable} and SiT~\cite{ma2024sit}; (b) Zoomed in results around SiT-B/2.}
  \label{fig:main}
\end{figure}

We adopt the optimized thresholds obtained in \cref{sec:schedule} and repeat the aforementioned experiment on a larger scale model SiT-B/2. The results in \cref{fig:main} show that our method consistently outperform both DiT and SiT (\cref{fig:main} (a)) and this phenomenon is consistent between different model sizes (\cref{fig:schedule} (a) for SiT-S/2, \cref{fig:main} (b) for SiT-B/2). Specifically, our method can get a better FID score (1.85 lower than SiT-B/2) with even less computation budget.

\begin{figure}[t]
  \centering
  \includegraphics[width=1.0\linewidth]{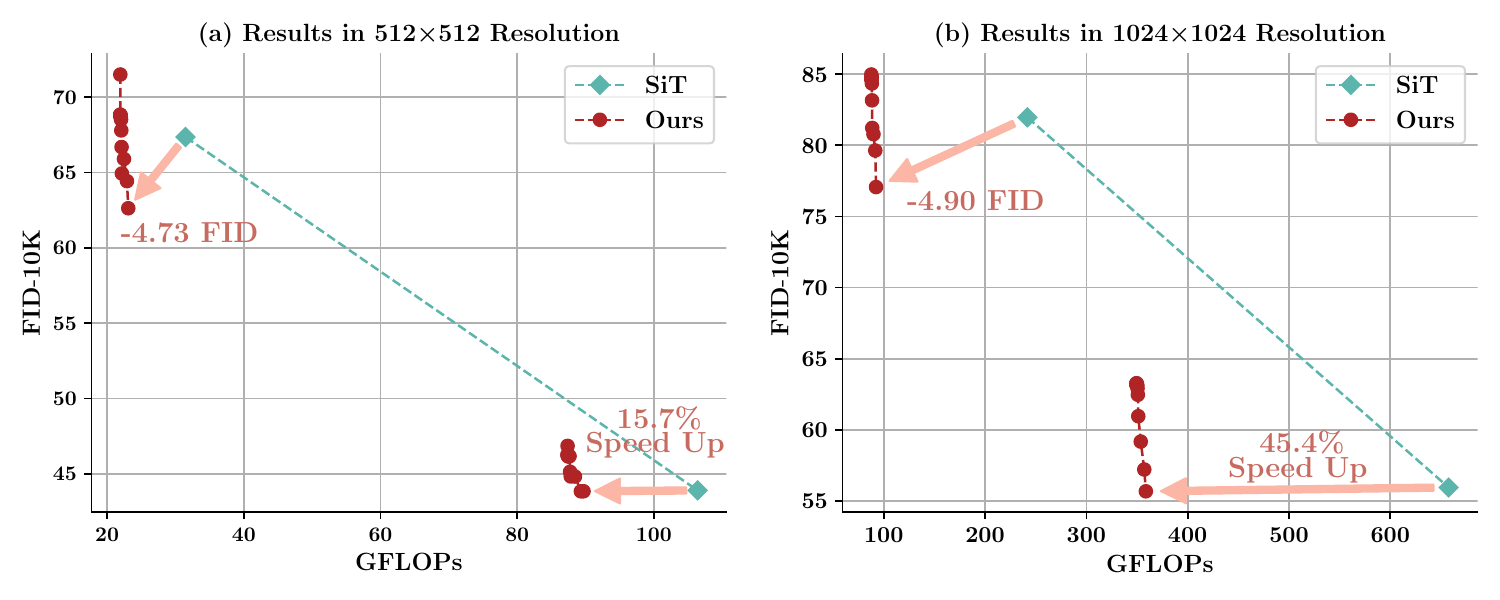}
  \caption{High resolution image generation results.}
  \label{fig:hr}
\end{figure}

\begin{table}
  \caption{Benchmarking class-conditional image generation on ImageNet 256$\times{}$256. }
  \label{tab:sota}
  \centering
  \begin{tabular}{cccccc}
    \toprule
    Model & FID$\downarrow$ & sFID$\downarrow$ & IS$\uparrow$ & Precision$\uparrow$ & Recall$\uparrow$ \\
    \midrule
    BigGAN-deep\cite{brock2019large} & 6.95 & 7.36 & 171.4 & \underline{\textbf{0.87}} & 0.28 \\
    StyleGAN-XL\cite{sauer2022styleganxl} & 2.30 & \underline{\textbf{4.02}} & 265.12 & 0.78 & 0.53 \\
    Mask-GIT\cite{chang2022maskgit} & 6.18 & - & 182.1 & - & -\\
    ADM\cite{dhariwal2021diffusion} & 10.94 & 6.02 & 100.98 & 0.69 & \underline{\textbf{0.63}} \\
    ADM-G, ADM-U & 3.94 & 6.14 & 215.84 & \underline{0.83} & 0.53 \\
    CDM\cite{ho2021cascaded} & 4.88 & - & 158.71 & - & - \\
    RIN\cite{jabri2023scalable} & 3.42 & - & 182.0 & - & - \\
    Simple Diffusion(U-Net)\cite{hoogeboom2023simple} & 3.76 & - & 171.6 & - & -\\
    Simple Diffusion(U-ViT, L) & 2.77 & - & 211.8 & - & - \\
    VDM++\cite{kingma2023understanding} & 2.12 & - & 267.7 & - & -\\
    \midrule
    DiT-XL$_{\text{(cfg = 1.5)}}$~\cite{peebles2023scalable} & 2.27 & 4.60 & \underline{\textbf{278.24}} & \underline{0.83} & 0.57 \\
    SiT-XL$_{\text{(cfg = 1.5)}}$~\cite{ma2024sit}   & \underline{2.06} & 4.50 &  270.27 & 0.82 & 0.59\\
    \midrule
    Ours$_{\text{(cfg = 1.5)}}$   & \underline{\textbf{2.01}} & \underline{4.49} &  \underline{271.04} & 0.82 & \underline{0.60} \\
  \bottomrule
  \end{tabular}
\end{table}

We further conduct experiment on generating higher resolution images. The $512 \times 512$ resolution models are finetuned from $256 \times 256$ models with a global batch size of $64$ for $400$K iterations, while $1024 \times 1024$ models are finetuned from $512 \times 512$ counterparts with a global batch size of $16$ for $400$K iterations. For testing $512 \times 512$ resolution models, we generate 10K images with our model and compute the FID with 512 resolution reference batch obtained from \texttt{guided-diffusion}\footnote{https://github.com/openai/guided-diffusion/tree/main/evaluations}. For $1024 \times 1024$ models, we randomly select 10K images from ImageNet validation set, resize them into $1024^2$ resolution, and compute FID ( with \texttt{clean-fid}\footnote{https://github.com/GaParmar/clean-fid} toolkit) with 10K images sampled by our model.

The high-resolution results is illustrated in \cref{fig:hr}, where we can find that: (1) the proposed method can still achieve better generated image quality  (\eg, for SiT-S/2, $-4.90$ FID for $1024^2$) with far fewer FLOPs, and (2) the speedup is even more significant as the image resolution increases (\eg, for SiT-B/2, the speed-up increase from 15.7\% in $512^2$ resolution to 45.4\% in $1024^2$ resolution). This is because as the image resolution grows, the sequence length the attention operation needs to process also increases. At this point, the superiority of the linear complexity in our method becomes far more prominent compared to standard attention, which has quadratic complexity w.r.t the sequence length.

\subsection{Comparsion with State-of-the-art}\label{sec:sota}

We compare our method against state-of-the-art class-conditional generative models with the highest complexity SiT-XL/2 model endowed with our method. We replace the first four self-attention layers with the proposed attention with mediator tokens, and finetune the modified model for 400K iterations. The results reported in \cref{tab:sota} illustrate that when using classifier-free guidance (cfg=$1.5$), following the practice in DiT and SiT, our method outperforms all the prior diffusion models, achieving a remarkable FID-50K of $2.01$.

\subsection{Ablation Studies}\label{sec:ablation}

\begin{table}
  \caption{Effectiveness of static mediator tokens. $n$ is the mediator tokens number. }
  \label{tab:abl_redu}
  \centering
  \begin{tabular}{l|c|ccccc}
    \toprule
    Model & FLOPs(G) & FID ($\downarrow$) & sFID ($\downarrow$) & IS ($\uparrow$) & Precision ($\uparrow$) & Recall ($\uparrow$) \\
    \midrule
    SiT-S/2 (baseline) & 6.06 & 58.61 & 9.25 & 24.31 & 0.41 & 0.59\\
    \midrule
    $r=0.875$ & 5.91 & 58.98 & 9.10 & 24.13 & 0.40 & 0.60 \\
    $r=0.750$ & 5.76 & 59.18 & 9.26 & 24.03 & 0.39 & 0.59 \\
    $r=0.625$ & 5.61 & 60.30 & 9.58 & 23.74 & 0.39 & 0.59 \\
    $r=0.500$ & 5.46 & 60.02 & 9.43 & 24.01 & 0.40 & 0.57 \\
    \midrule
    Ours ($n=64$) & 5.78 & 53.57 & 9.01 & 27.26 & 0.43 & 0.61 \\
  \bottomrule
  \end{tabular}
\end{table}

\subsubsection{Comparison with vanilla Q-K compression.}

In order to verify that the proposed mediator token method is an effective way to leverage the query-key interaction redundancy, we design experiments where queries and keys are reduced in a simpler way. Specifically, in each self-attention layer of the SiT model, we modify the $\mathbf{W_q}$ and $\mathbf{W_k}$ linear projections from $\mathbb{R}^{C\times{}C}$ to $\mathbb{R}^{C\times{}rC}$ (where $r < 1$) dimensions. In this way, queries and keys also interact in a compressed space. We train this model with the same training recipe as SiT. The results in \cref{tab:abl_redu} show that although directly reducing the hidden dimension of queries and keys can save computation cost, the generated image quality drops dramatically. In contrast, the proposed method can increase the generated image quality as well as reduce the inference cost, verifying that our method is an effective way to leverage the redundancy in diffusion transformers.

\subsection{Visualization Results}\label{sec:vis}

\begin{figure}[t]
  \centering
  \includegraphics[width=1.0\linewidth]{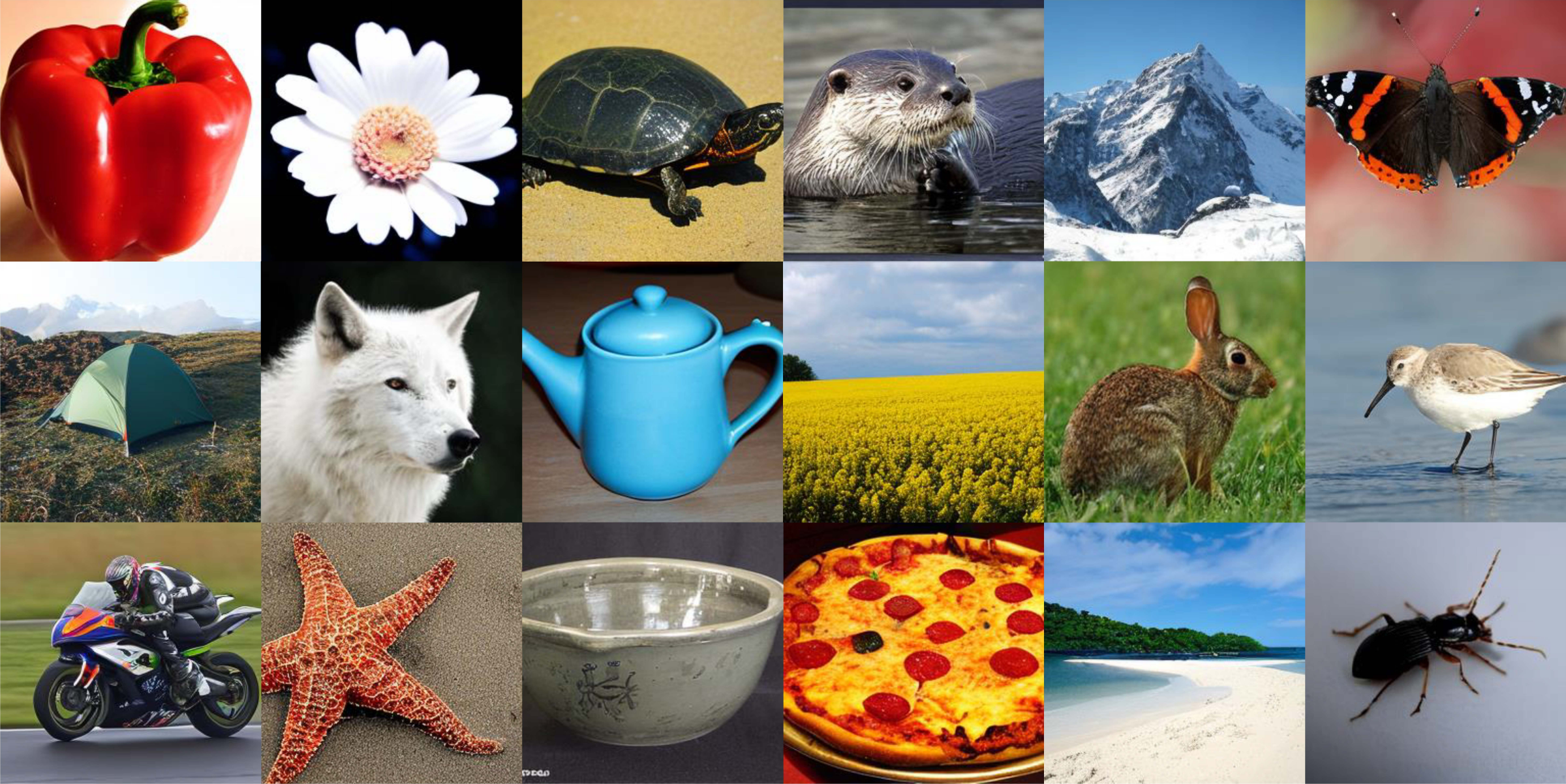}
  \caption{Sampled images by SiT-XL/2 models endowed with our method trained on ImageNet 256$\times{}$256 resolution with cfg$=$4.0.}
  \label{fig:vis}
\end{figure}

In order to verify the proposed time step-wise dynamic mediator token adjusting token mechanism does not achieve a better numerical result by over-fitting the FID-50K metric, we visualize the sample images using the largest SiT-XL/2 based model. Following the common practice in the DiT~\cite{peebles2023scalable} and the SiT~\cite{ma2024sit}, we set the classifier-free guidance as $4.0$ to sample the images. The sampled results are visualized in \cref{fig:vis}, from which we can find that our method not only can achieve lower FID metric but also can generate high-quality images.

\section{Conclusion}
\label{sec:conclusion}

This paper proposed a novel diffusion transformer architecture in which an extra group of mediator tokens interact with the query tokens and key tokens separately, compressing the redundant query-key interaction during the denoising generation process. The number of mediator tokens adjusts across different denoising time steps conditioned on the difference between every two adjacent latent features in a simple-wise dynamic manner. Extensive quantitative experiments and qualitative generated results demonstrate the effectiveness of our method in alleviating attention redundancy and improving the generated image quality. Our method also reduces the computation complexity in the attention model since the proposed mechanism makes the attention operation have linear complexity with regard to the image token length.

\section*{Acknowledgements}
This work is supported in part by the National Natural Science Foundation of China under Grants 62321005 and 42327901.


%
%
\bibliographystyle{splncs04}
\bibliography{main}

\input
\end{document}